\documentclass{article}

\usepackage{arxiv}

\usepackage[utf8]{inputenc} 
\usepackage[T1]{fontenc}    
\usepackage{hyperref}       
\usepackage{url}            
\usepackage{booktabs}       
\usepackage{amsfonts}       
\usepackage{nicefrac}       
\usepackage{microtype}      
\usepackage{lipsum}
\usepackage{graphicx}
\graphicspath{ {./images/} }
\usepackage{amsmath}
\usepackage{booktabs}
\usepackage{multirow}
\usepackage{graphicx}
\usepackage[table]{xcolor}
\usepackage{multirow}
\usepackage{algorithm}
\usepackage{algorithmic}
\usepackage{amssymb}

\title{Raptor: Rejection-aware Adaptive Tree based on Speculative Decoding}

\author{
 Qishu Hu \\
  School of Automation\\
  Beijing Institute of Technology\\
  Beijing, China 100081 \\
  \texttt{3220251165@bit.edu.cn} \\
   \And
 Haozhao Wang \\
  School of Computer Science and Technology\\
  Huazhong University of Science and Technology\\
  Wuhan, China 430074‌ \\
  \texttt{hz\_wang@hust.edu.cn} \\
  \And
 Zicong Hong \\
  Distributed Computing Lab\\
  École Polytechnique Fédérale de Lausanne\\
  Lausanne, Switzerland 1015 \\
  \texttt{zicong.hong@epfl.ch} \\
  \And
 Yufeng Zhan\thanks{Corresponding author.} \\
  School of Automation\\
  Beijing Institute of Technology\\
  Beijing, China 100081 \\
  \texttt{yu-feng.zhan@bit.edu.cn} \\
  \And
 Yuanqing Xia \\
  School of Automation\\
  Beijing Institute of Technology\\
  Beijing, China 100081 \\
  \texttt{xia\_yuanqing@bit.edu.cn} \\
}

\begin{document}
\maketitle
\begin{abstract}
Tree-structured speculative decoding accelerates vision-language models (VLMs) by verifying multiple candidate continuations in parallel. However, long visual contexts increase the cost of speculative computation, making candidate coverage and redundant expansion important considerations beyond tree topology adaptation. We propose \textbf{Raptor}, a training-free framework that uses rejection feedback to jointly regulate candidate coverage and drafting effort. Our key insight is that acceptance length alone cannot distinguish insufficient candidate coverage from excessive deep expansion, and these conditions require different allocation decisions. Raptor therefore extracts complementary rejection-aware signals to characterize speculative progress, frontier coverage, and redundant expansion. These signals jointly guide bounded, momentum-smoothed updates of tree depth, branching width, and drafting budget, together with execution control that limits unnecessary drafting and progressively restores expansion when appropriate. Extensive experiments with two VLMs across seven multimodal benchmarks demonstrate the effectiveness of Raptor, achieving up to 2.39× average speedup over autoregressive decoding.
\end{abstract}


\begin{figure*}[t]
\centering
\includegraphics[width=1.0\textwidth]{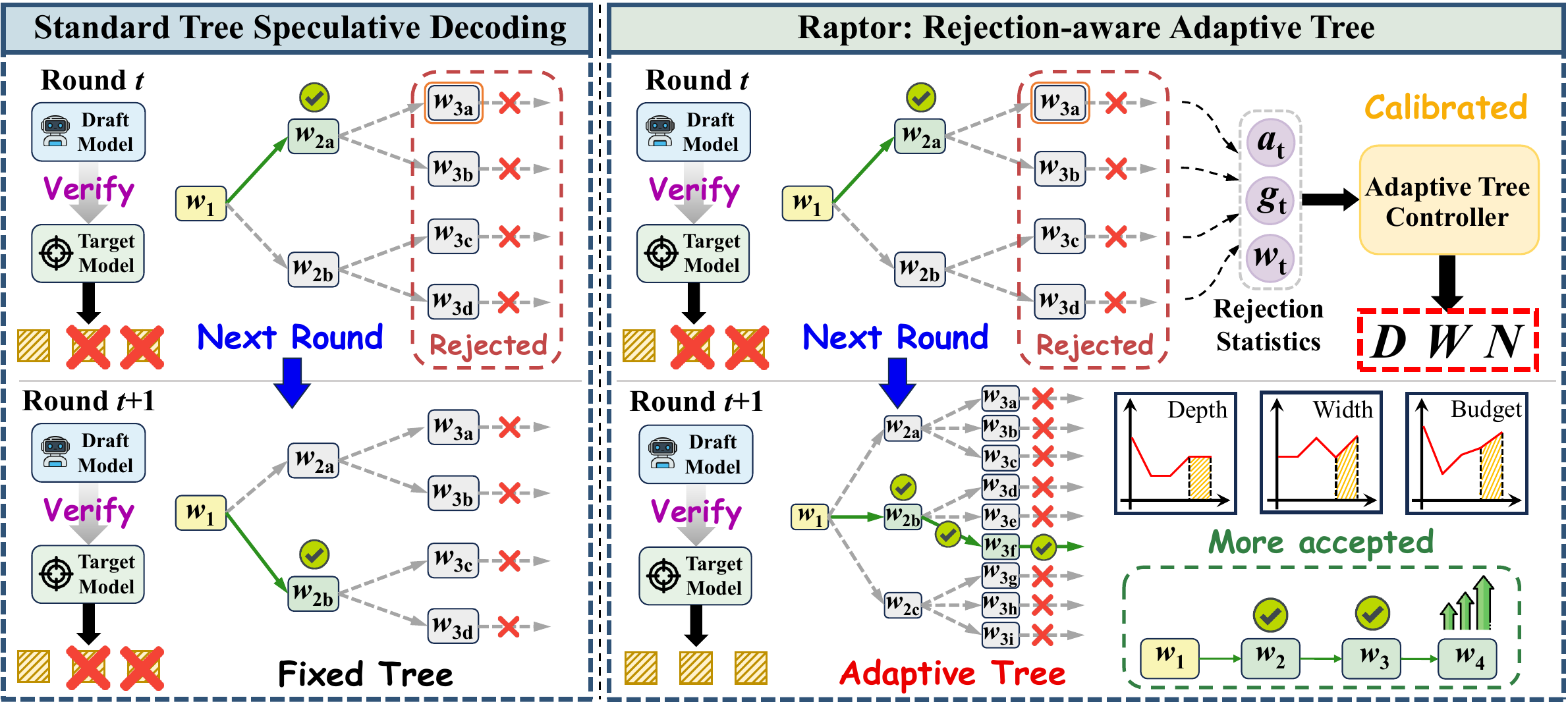} 
\caption{\textbf{Overview of Rejection-aware Raptor.} Conventional tree-structured speculative decoding relies on fixed draft tree configurations, where rejected branches are discarded without contributing to future tree construction. Raptor reuses rejection information as feedback, calibrates the next draft tree through an adaptive controller, and dynamically balances candidate coverage and speculative waste across decoding rounds.}
\label{fig1}
\end{figure*}

\section{Introduction}
\label{sec:intro}
Vision-language models (VLMs) have achieved strong performance in visual question answering\cite{liu2023visual, liu2024improved}, image captioning\cite{clark2026molmo2, xiao2024florence}, and multimodal reasoning\cite{chen2024expanding}. Recent models further extend these capabilities to video understanding and grounding\cite{deitke2025molmo}. However, their token-by-token autoregressive generation\cite{brown2020language, touvron2023llamaopenefficientfoundation} remains costly. Beyond sequential token generation, high-resolution visual inputs can substantially enlarge the context processed by the Transformer-based language backbone\cite{vaswani2017attention, wu2025qwen}, increasing both attention computation and the memory required to store cached keys and values during decoding\cite{kwon2023efficient}. Speculative decoding must therefore balance the benefit of reducing target-model iterations against the additional cost of drafting and verifying candidate tokens\cite{chen2024sequoia, sun2023spectr}. In this setting, speculative computation is useful only when the accepted progress justifies its execution cost. This makes efficient candidate allocation particularly important for VLM inference, where multimodal contexts increase the cost of both candidate construction and target-model verification.

Speculative decoding reduces target-model iterations by drafting candidate tokens\cite{chen2023accelerating, leviathan2023fast} and verifying them in parallel\cite{du2024glide}. Tree-based variants further improve candidate coverage by exploring multiple continuations\cite{svirschevski2024specexec}. Adaptive tree construction has already been investigated: EAGLE-2 uses draft confidence to select context-dependent trees\cite{li-etal-2024-eagle}, TALON allocates a fixed node budget through confidence-aware expansion\cite{liu2026talon}, and SAGE combines entropy-guided adaptation with acceptance history\cite{tong2026sage}. These uncertainty-driven approaches establish the value of adaptive trees, but confidence and entropy do not explicitly quantify which target-preferred alternatives are missing or how much expansion extends beyond accepted progress. We instead use rejection feedback, defined as target-verification evidence about accepted progress, missing probability mass at the rejection frontier, and draft nodes beyond the accepted prefix. This feedback links candidate coverage and redundant expansion to subsequent computation allocation, extending adaptation beyond confidence estimates and acceptance summaries.

Our analysis highlights rejection feedback as a basis for jointly assessing candidate coverage and redundant expansion to guide computation allocation. Turning this evidence into effective allocation decisions presents three challenges. First, short accepted prefixes do not specify how to reallocate computation: missing frontier probability mass motivates broader exploration, whereas excessive expansion beyond the accepted prefix calls for reduced depth and computation. These conditions can coexist, creating competing demands. Second, depth, width, and node budget are interdependent. Under a fixed budget, wider branching leaves fewer nodes for deeper paths, while greater depth can increase serial drafting cost. Independent adjustments may therefore compromise coverage or efficiency. Third, verification evidence is incomplete and varies across rounds. Full acceptance of the selected path leaves further acceptance unknown, so path exhaustion should not trigger the same restriction as observed rejection. Abrupt responses to individual rounds can also destabilize adaptation. Effective control must therefore reconcile coverage and waste, coordinate tree parameters, and distinguish rejection from path exhaustion while smoothing temporal fluctuations.

Accordingly, we propose Raptor, a training-free framework for rejection-aware computation allocation in tree-structured speculative decoding, as illustrated in Figure~\ref{fig1}. Raptor summarizes each verification round through acceptance length, frontier gap, and deep waste, which describe achieved progress, missing local candidate coverage, and expansion beyond the accepted prefix. These measurements drive a bounded controller that jointly adjusts tree depth, branching width, and node budget through momentum-smoothed updates. An accompanying execution mechanism limits serial expansion after genuine shallow rejection and progressively relaxes this limit when the selected path is exhausted. Together, these mechanisms adapt candidate allocation and drafting effort while retaining the original drafter and target-verification process. By reusing verification statistics, Raptor requires no additional training or target-model forward pass and integrates with existing tree-structured speculative decoding frameworks without architectural changes.

Our contributions are summarized as follows:
\begin{itemize}
\item We reveal that acceptance length alone is insufficient to guide speculative computation allocation, and show how rejection feedback exposes complementary information about candidate coverage and redundant expansion.
\item We propose Raptor, a training-free framework that coordinates tree configuration and rejection-aware execution control to regulate speculative computation across rounds.
\item We evaluate Raptor on seven multimodal benchmarks with two representative VLMs, demonstrating consistent speedup gains across three tree-based speculative decoding frameworks and outperforming existing adaptive strategies in most evaluated settings.
\end{itemize}

\section{Related Work}
\label{sec:formatting}

\subsection{Speculative Decoding for Efficient Generation}

Speculative decoding accelerates autoregressive generation through a draft-and-verify paradigm. Specifically, a lightweight draft model first generates a sequence of candidate tokens, which are then verified in parallel by the target model. Given a draft sequence ${y_{1:K}^D}$, the target model accepts the longest prefix satisfying the verification criterion:
\begin{equation}
    \max \left\{ {l \mid y_{1:l}^D = y_{1:l}^T} \right\},
\end{equation}

\noindent where $y^D$ and $y^T$ denote the token sequences generated by draft model and target model, respectively. By reducing the number of expensive target-model decoding steps while preserving generation consistency, speculative decoding provides a lossless acceleration framework\cite{2023Speculative}. Various studies have explored alternative candidate generation mechanisms, including retrieval-based drafting\cite{2024REST}, prompt-based matching \cite{lee2025promtec}, and Jacobi decoding\cite{santilli2023accelerating, kou2024cllms}. Nevertheless, speculative decoding remains highly sensitive to draft quality, where each rejection may lead to discarding the corresponding speculative suffix.

\subsection{Tree-Structured Speculative Decoding.}

Tree-structured speculative decoding extends conventional linear speculation by allowing the draft model to explore multiple candidate continuations simultaneously. Instead of generating a single sequence, the draft model constructs a tree $\mathcal{T}$ containing multiple speculative paths. Let $L_m$ denote the length of the $m$-th candidate path and $M$ represent the number of speculative paths, these paths are jointly verified by the target model:
\begin{equation}
    \mathcal{T} = \left\{ {y_{1:{L_m}}^{\left( m \right)}} \right\}_{m = 1}^M.
\end{equation}

By increasing candidate diversity, tree-based methods reduce early rejection and improve the probability of accepting longer speculative trajectories\cite{ankner2024hydra, kim2024exploring}. SpecInfer\cite{miao2024specinfer} first demonstrates the effectiveness of tree-based verification for large model serving. Subsequently, Medusa\cite{cai2024medusa} introduces multiple prediction heads for parallel future token generation, while EAGLE\cite{li2024eagle} improves candidate quality through feature-level uncertainty modeling. In multimodal scenarios, ViSpec\cite{kang2026vispec} extends tree-structured speculative decoding to vision-language models by incorporating visual information into candidate generation\cite{gagrani2024speculative}. However, despite adapting trees to prediction uncertainty and acceptance behavior, existing approaches insufficiently exploit verification feedback to distinguish limited candidate coverage from redundant expansion. This distinction guides where to expand and how much computation to allocate beyond confidence- or acceptance-based topology adjustment. These limitations motivate us to investigate structural information in rejection outcomes and design coordinated feedback for subsequent tree adaptation, without additional model training or changes to the underlying model architectures.

\section{Method}
\label{sec:method}

This section presents Raptor, a training-free framework that uses rejection feedback to adapt speculative decoding. We first introduce complementary rejection signals that characterize accepted progress, frontier coverage, and redundant expansion (Sec. \ref{4.1}). We then describe a rejection-calibrated controller that maps these signals into coordinated adjustments of tree depth, branching width, and drafting budget (Sec. \ref{4.2}). Finally, we present momentum-based updates and discrete projection that translate these adjustments into valid configurations for subsequent tree construction (Sec. \ref{4.3}).

\subsection{Rejection Structure Representation}
\label{4.1}

\begin{figure*}[!t]
\centering
\includegraphics[width=1.0\textwidth]{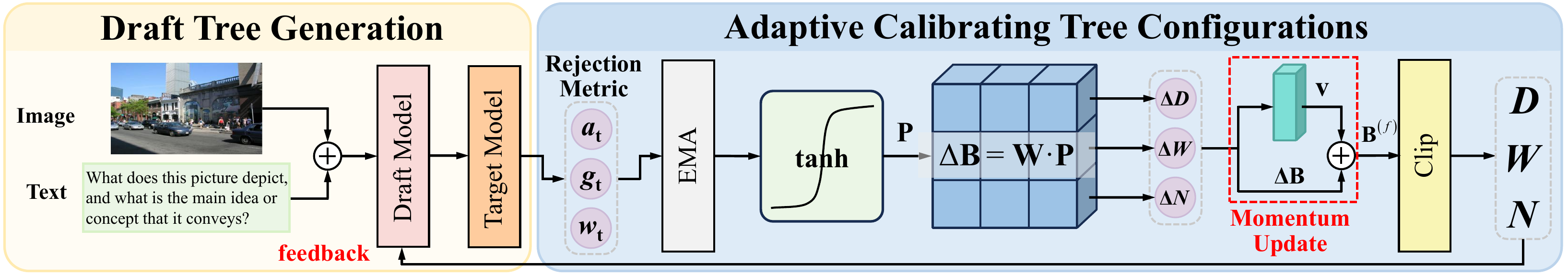}
\caption{\textbf{Detailed optimization workflow of the Raptor framework.} Given multimodal inputs, the draft model constructs speculative trees that are verified by the target model, from which three rejection-aware signals, including acceptance length, frontier gap, and deep waste, are extracted. These signals are converted into adaptive controls to update tree depth, branching width, and drafting budget through momentum-based regulation.}
\label{fig4}
\end{figure*}

Raptor uses verification feedback from the current decoding round to adapt the draft tree for the next round. As illustrated in Figure~\ref{fig4}, it first extracts three complementary rejection-aware statistics and then converts them into stable updates of tree depth, branching width, and total drafting budget.

Acceptance length measures the progress achieved by the selected draft path in the current verification round, reflecting draft–target agreement along its accepted prefix. Given the candidate path \textit{P}, it is defined as
\begin{equation}
    {a_t} = \mathop {\max }\limits_{P \in {\cal P}} \sum\limits_{i = 1}^{\left| P \right|} {\prod\limits_{j = 1}^i {\textbf{1}(y_j^{draft} = y_j^{t\arg et})} },
\end{equation}

\noindent where ${\cal P}$ denotes the set of candidate paths, and $\textbf{1}\left(  \cdot  \right)$ is the binary acceptance indicator function, returning 1 when the draft token agrees with the target model prediction and 0 otherwise.

Although acceptance length reflects progress along the selected path, it does not quantify candidate coverage at the rejection frontier. We therefore introduce frontier gap, a distribution-aware metric that measures the target probability mass not covered by candidate children at that frontier. 

Specifically, let ${p_t}\left( {y|{x_r}} \right)$ denote the target model probability distribution at the first rejected position ${x_r}$. Given the candidate token set ${\cal Y}$ explored by the current draft tree, the Frontier Gap $g_t$ is defined as
\begin{equation}
    {g_t} = 1 - \sum\limits_{y \in {\cal Y}} {{p_t}\left( {y|{x_r}} \right)}. 
\end{equation}

\noindent The target probabilities are obtained from the existing verification pass. When the selected path is exhausted, no rejection frontier is observed, and $g_t$ is set to zero.

While frontier gap measures missing candidate coverage, expansion beyond the accepted prefix introduces speculative work that does not contribute to the current round’s progress. We therefore introduce deep waste to characterize such redundant expansion. It measures the fraction of non-root draft nodes whose depth exceeds the accepted draft length and is defined as
\begin{equation}
    {w_t} = \frac{1}{N}\sum\limits_i {\textbf{1}\left( {{d_i} > {a_t}} \right)},
\end{equation}

\noindent where \(N\) denotes the number of draft nodes and \(d_i\) is the depth of the \(i\)-th node.

\subsection{Rejection-Calibrated Adaptive Controller}
\label{4.2}

The rejection-aware statistics introduced before provide complementary information about the current speculative decoding state. However, due to the fact that these original signals have different units, meanings and optimization directions, directly using them for tree structure adjustment poses challenges. To address this issue, we propose a rejection-calibrated adaptive controller that transforms heterogeneous feedback signals into unified control signals and regulates three fundamental draft tree parameters.

\subsubsection{Unified Saturated Pressure Mapping}

The three rejection-aware signals have different scales and control directions. Directly applying linear scaling may cause unstable updates, so Raptor first applies exponential moving average (EMA) smoothing to suppress short-term fluctuations:
\begin{equation}
    {\bar{r} _t} = \beta {\bar{r}_{t - 1}} + \left( {1 - \beta } \right){r_t},{\quad \rm{     }}{r_t} \in \left\{ {{a_t},{g_t},{w_t}} \right\},
\end{equation}

\noindent where $\beta$ represents the EMA coefficient, its value reflects the overall trend of the recent indicators. The $r_t$ denotes the current observed rejection statistic.

EMA reduces short-term fluctuation caused by individual decoding steps and provides a more stable estimation of the current generation state. Raptor then maps the deviation of each smoothed statistic from its operating point into a bounded pressure:
\begin{equation}
    {P_i} = \tanh \left( {\frac{{{\Delta _i}}}{{{\sigma _i}}}} \right),
\end{equation}

\noindent where ${P_i}$ denotes the pressure corresponding to the signal \textit{i}, ${\Delta _i}$ indicates the deviation between the corresponding feedback statistics and the expected working point, ${\sigma _i}$ is a normalization factor which controls the sensitivity of the response.

The hyperbolic tangent maps heterogeneous signals into a shared bounded space and prevents extreme observations from causing abrupt configuration changes. The resulting pressure vector is given by
\begin{equation}
    {{\bf{P}}_t} = {\left[ {{P_t^a}{\text{ }}{P_t^g}{\text{ }}{P_t^w}} \right]^{\rm{T}}}.
\end{equation}

\subsubsection{Pressure-based Budget Regulation}

After obtaining the unified pressure representation, Raptor converts rejection feedback into adaptive adjustments of draft tree configurations. The bounded pressures are combined to jointly regulate tree depth, branching width, and drafting budget. We represent the current tree configuration as:
\begin{equation}
    {{\bf{B}}_t} = {\left[ {{D_t}{\text{ }}{W_t}{\text{ }}{N_t}} \right]^{\rm{T}}},
\end{equation}

\noindent where ${D_t}$, ${W_t}$ and ${N_t}$ denote tree depth, branching width, and total draft token budget, respectively.

The adjustment of tree configuration is computed through a unified controller:
\begin{equation}
    \Delta {{\mathbf{B}}_t} = {\mathbf{W}} \cdot {{\mathbf{P}}_t},{\text{   }}{\mathbf{W}} \in {\mathbb{R}^{3 \times 3}},
\end{equation}

\noindent where \textbf{W} is the pressure-to-budget transformation matrix. Each row of \textbf{W} determines how different rejection pressures influence one specific tree parameter, while each column represents the contribution of one rejection signal. Therefore, it combines complementary feedback into coordinated updates. Its coefficients and calibration procedure are detailed in Appendix A.2.

Compared with independent heuristic rules, this unified formulation provides a more compact and interpretable mechanism for adaptive tree control. It enables Raptor to dynamically allocate speculative computation according to the current verification state while maintaining a consistent optimization objective across different tree parameters.

\subsection{Momentum-based Continuous Tree Update}
\label{4.3}

Although the pressure-based controller provides adaptive adjustments for draft tree configurations, directly applying the instantaneous update $\Delta {{\bf{B}}_t}$ at each decoding iteration may introduce unstable oscillations. Verification feedback can vary across consecutive rounds even under greedy decoding as the accepted prefix and candidate tree change. Although EMA smooths the observed statistics, directly applying the resulting configuration increments may still cause abrupt adjustments. Raptor therefore maintains a momentum-smoothed update velocity that combines historical updates with the current feedback-driven increment. The continuous configuration is updated as follows:
\begin{equation}
\begin{array}{c}
{{\bf{v}}_t} = \mu {{\bf{v}}_{t - 1}} + \left( {1 - \mu } \right)\Delta {{\bf{B}}_t},\\[1.5ex]
{\bf{B}}_{t + 1}^{\left( f \right)} = {\bf{B}}_t^{\left( f \right)} + \eta {{\bf{v}}_t},
\end{array}
\end{equation}

\noindent where $\bf{v}_t$ denotes the momentum-smoothed update velocity, $\mu$ is the momentum coefficient, ${\bf{B}}_t^{\left( f \right)}$ denotes the continuous-valued tree configuration before discrete projection, and $\eta$ controls the update step size.

By accumulating historical update information, the momentum mechanism prevents the controller from overreacting to individual decoding steps and produces smoother adaptation trajectories. Meanwhile, the step-size factor $\eta$ provides additional control over the adaptation magnitude, enabling gradual adjustment of tree configurations according to long-term verification feedback. 

After obtaining the continuous configuration ${\bf{B}}_{t + 1}^{\left( f \right)}$, Raptor applies a discrete projection operation:
\begin{equation}
\begin{array}{c}
  {{\mathbf{B}}_{t + 1}} = \text{clip}\left( {\text{round}\left( {{\mathbf{B}}_{t + 1}^{\left( f \right)}} \right)} \right),\\[1.5ex]
  {\mathbf{B}}_{t + 1}^{\left( f \right)} \in {\mathbb{R}^3},{\text{ }}{{\mathbf{B}}_{t + 1}} \in {\mathbb{Z}^3}.
\end{array} 
\end{equation}

The continuous state is constrained by parameter bounds and candidate capacity before rounding, followed by an integer-capacity check. The node-budget channel uses twice the base step size.  The resulting discrete configuration specifies the next tree’s allocation limits, while verification feedback further regulates the actual expansion process and preserves structural validity.

Besides, to control drafting effort within this configuration, Raptor also distinguishes genuine shallow rejection from path exhaustion. A shallow rejection requests a reduced expansion range for the next round, whereas path exhaustion progressively relaxes an existing limit because further acceptance remains unobserved. Expansion may continue beyond the preferred limit when the candidate pool is insufficient, subject to architectural constraints. Together, configuration adaptation and execution control enable Raptor to continuously adapt to changing generation difficulty without additional training or modifications to the underlying model architectures. Full update rules and drafter-specific integration are provided in Appendix A.3–A.4.

\definecolor{raptorgreen}{RGB}{0,128,0}
\definecolor{speedred}{RGB}{200,0,0}
\definecolor{speedgray}{RGB}{80,80,80}
\definecolor{raptorblue}{RGB}{226,239,250}

\newcommand{\speedannotation}[3]{%
  \hspace{0.06em}%
  \raisebox{-0.35ex}{%
    \scalebox{0.62}{%
      \normalfont%
      \textcolor{#1}{%
        \begingroup%
        \mathsurround=0pt%
        \ensuremath{\mathord{#2}}%
        \kern-0.08em%
        #3\%%
        \endgroup%
      }%
    }%
  }%
}

\newcommand{\speedup}[2]{%
  #1\speedannotation{raptorgreen}{\uparrow}{#2}%
}

\newcommand{\speeddown}[2]{%
  #1\speedannotation{speedred}{\downarrow}{#2}%
}

\newcommand{\speedsame}[2]{%
  #1\speedannotation{speedgray}{\rightarrow}{#2}%
}

\newcommand{\raptorspeed}[2]{%
  \textbf{#1}%
  \speedannotation{raptorgreen}{\uparrow}{\textbf{#2}}%
}

\begin{table*}[!t]
\centering
\caption{Comparison between Raptor and existing adaptive tree strategies across VLMs, speculative decoding frameworks, and multimodal benchmarks, with original speculative decoding configurations included as baselines. Speedups (Spd.) are measured relative to autoregressive decoding, and the highest speedup is highlighted in bold.}
\label{table1}
\setlength{\tabcolsep}{2.6pt}
\renewcommand{\arraystretch}{1.08}
\scriptsize

\edef\savedaboverulesep{\the\aboverulesep}
\edef\savedbelowrulesep{\the\belowrulesep}

\setlength{\aboverulesep}{0.2ex}
\setlength{\belowrulesep}{0.2ex}

\resizebox{\textwidth}{!}{%
\begin{tabular}{c|cccccccccccccc|cc}
\toprule
\multirow[c]{2}{*}[-\dimexpr
  \aboverulesep/2+\belowrulesep/2+\cmidrulewidth/2\relax]{Method} &
\multicolumn{2}{c}{COCO Caps} &
\multicolumn{2}{c}{MME} &
\multicolumn{2}{c}{MM-Vet} &
\multicolumn{2}{c}{SQA} &
\multicolumn{2}{c}{TextVQA} &
\multicolumn{2}{c}{VizWiz} &
\multicolumn{2}{c}{MMMU} &
\multicolumn{2}{|c}{Avg.} \\
\cmidrule(lr){2-3}\cmidrule(lr){4-5}
\cmidrule(lr){6-7}\cmidrule(lr){8-9}
\cmidrule(lr){10-11}\cmidrule(lr){12-13}
\cmidrule(lr){14-15}\cmidrule(lr){16-17}
& $\tau$ & Spd. & $\tau$ & Spd.
& $\tau$ & Spd. & $\tau$ & Spd.
& $\tau$ & Spd. & $\tau$ & Spd.
& $\tau$ & Spd. & $\tau$ & Spd. \\
\noalign{%
  \global\aboverulesep=\savedaboverulesep\relax
  \global\belowrulesep=\savedbelowrulesep\relax
}
\midrule
\multicolumn{17}{c}{\textit{LLaVA-1.6}} \\
\specialrule{\lightrulewidth}{\aboverulesep}{0.2ex}
Medusa
& 0.27 & 1.02x & 0.46 & 1.20x
& 0.58 & 1.74x & 0.37 & 1.09x
& 0.60 & 1.39x & 0.32 & 1.04x
& 0.67 & 1.53x & 0.47 & 1.29x \\
\specialrule{\lightrulewidth}{0.2ex}{\belowrulesep}
+SAGE
& 0.06 & \speeddown{0.91x}{10.8} & 0.06 & \speeddown{0.90x}{25.0}
& 0.20 & \speeddown{1.39x}{20.1} & 0.09 & \speeddown{0.96x}{11.9}
& 0.32 & \speeddown{1.14x}{18.0} & 0.15 & \speeddown{0.96x}{7.7}
& 0.25 & \speeddown{1.13x}{26.1} & 0.16 & \speeddown{1.06x}{17.8} \\
+TALON
& 0.56 & \speedup{1.22x}{19.6} & 0.54 & \speedup{1.24x}{3.3}
& 0.73 & \speedup{2.16x}{24.1} & 0.63 & \speedup{1.14x}{4.6}
& 0.95 & \speedup{1.67x}{20.1} & 0.59 & \speedup{1.18x}{13.5}
& 0.83 & \speedup{1.55x}{1.3} & 0.69 & \speedup{1.45x}{12.4} \\
\rowcolor{raptorblue}
+Raptor
& 0.65 & \raptorspeed{1.35x}{32.4} & 0.47 & \raptorspeed{1.36x}{13.3}
& 0.86 & \raptorspeed{2.45x}{40.8} & 0.87 & \raptorspeed{1.33x}{22.0}
& 1.10 & \raptorspeed{1.81x}{30.2} & 0.75 & \raptorspeed{1.40x}{34.6}
& 1.06 & \raptorspeed{1.70x}{11.1} & 0.82 & \raptorspeed{1.63x}{26.4} \\
\specialrule{\lightrulewidth}{\aboverulesep}{0.2ex}
EAGLE
& 0.58 & 1.02x & 0.11 & 0.86x
& 0.08 & 1.10x & 0.40 & 0.92x
& 0.48 & 0.97x & 0.50 & 0.95x
& 0.19 & 0.89x & 0.33 & 0.96x \\
\specialrule{\lightrulewidth}{0.2ex}{\belowrulesep}
+SAGE
& 0.52 & \speedup{1.03x}{1.0} & 0.11 & \speeddown{0.81x}{5.8}
& 0.07 & \speeddown{1.03x}{6.4} & 0.36 & \speedup{0.94x}{2.2}
& 0.43 & \speedup{0.98x}{1.0} & 0.44 & \speeddown{0.93x}{2.1}
& 0.38 & \speedup{0.94x}{5.6} & 0.33 & \speeddown{0.95x}{1.0} \\
+TALON
& 0.61 & \speeddown{0.99x}{2.9} & 0.15 & \speeddown{0.77x}{10.5}
& 0.09 & \speeddown{1.06x}{3.6} & 0.46 & \speedup{0.95x}{3.3}
& 0.51 & \speedsame{0.97x}{0.0} & 0.50 & \speedup{0.99x}{4.2}
& 0.45 & \speedsame{0.89x}{0.0} & 0.40 & \speeddown{0.95x}{1.0} \\
\rowcolor{raptorblue}
+Raptor
& 0.62 & \raptorspeed{1.21x}{18.6} & 0.12 & \raptorspeed{0.94x}{9.3}
& 0.08 & \raptorspeed{1.44x}{30.9} & 0.45 & \raptorspeed{1.06x}{15.2}
& 0.51 & \raptorspeed{1.14x}{17.5} & 0.51 & \raptorspeed{1.15x}{21.1}
& 0.45 & \raptorspeed{1.02x}{14.6} & 0.39 & \raptorspeed{1.14x}{18.8} \\
\specialrule{\lightrulewidth}{\aboverulesep}{0.2ex}
ViSpec
& 3.30 & 2.38x & 2.57 & 1.49x
& 2.63 & 2.66x & 2.98 & 1.57x
& 3.61 & 2.38x & 3.25 & 2.11x
& 3.44 & 2.07x & 3.11 & 2.09x \\
\specialrule{\lightrulewidth}{0.2ex}{\belowrulesep}
+SAGE
& 2.38 & \speeddown{1.93x}{18.9} & 1.62 & \speeddown{1.45x}{2.7}
& 1.96 & \speeddown{2.26x}{15.0} & 1.89 & \speeddown{1.45x}{7.6}
& 2.05 & \speeddown{1.79x}{24.8} & 2.00 & \speeddown{1.63x}{22.7}
& 1.94 & \speeddown{1.68x}{18.8} & 1.98 & \speeddown{1.74x}{16.7} \\
+TALON
& 3.33 & \speedup{2.49x}{4.6} & 2.64 & \speedup{1.72x}{15.4}
& 2.76 & \speedup{3.09x}{16.2} & 3.09 & \speedup{1.60x}{1.9}
& 3.18 & \speedup{2.39x}{0.4} & 2.78 & \speeddown{2.10x}{0.5}
& 2.98 & \speeddown{2.06x}{0.5} & 2.97 & \speedup{2.21x}{5.7} \\
\rowcolor{raptorblue}
+Raptor
& 3.64 & \raptorspeed{2.71x}{13.9} & 2.73 & \raptorspeed{1.79x}{20.1}
& 2.81 & \raptorspeed{3.21x}{20.7} & 2.81 & \raptorspeed{1.92x}{22.3}
& 3.09 & \raptorspeed{2.47x}{3.8} & 2.96 & \raptorspeed{2.27x}{7.6}
& 2.96 & \raptorspeed{2.33x}{12.6} & 3.00 & \raptorspeed{2.39x}{14.4} \\
\midrule
\multicolumn{17}{c}{\textit{Qwen2.5-VL}} \\
\specialrule{\lightrulewidth}{\aboverulesep}{0.2ex}
Medusa
& 0.13 & 0.68x & 0.13 & 0.74x
& 0.16 & 0.68x & 0.15 & 0.68x
& 0.17 & 0.75x & 0.17 & 0.74x
& 0.24 & 0.76x & 0.16 & 0.72x \\
\specialrule{\lightrulewidth}{0.2ex}{\belowrulesep}
+SAGE
& 0.09 & \speedup{0.73x}{7.4} & 0.06 & \speedup{0.81x}{9.5}
& 0.07 & \speedup{0.74x}{8.8} & 0.05 & \speedup{0.80x}{17.6}
& 0.08 & \speedsame{0.75x}{0.0} & 0.11 & \speedup{0.78x}{5.4}
& 0.05 & \speeddown{0.73x}{3.9} & 0.07 & \speedup{0.76x}{5.6} \\
+TALON
& 0.18 & \speedup{0.82x}{20.6} & 0.18 & \speedup{0.86x}{16.2}
& 0.22 & \speedup{0.84x}{23.5} & 0.22 & \speedup{0.85x}{25.0}
& 0.21 & \speedup{0.86x}{14.7} & 0.22 & \speedup{0.92x}{24.3}
& 0.32 & \speedup{0.91x}{19.7} & 0.22 & \speedup{0.87x}{20.8} \\
\rowcolor{raptorblue}
+Raptor
& 0.20 & \raptorspeed{0.91x}{33.8} & 0.19 & \raptorspeed{0.90x}{21.6}
& 0.26 & \raptorspeed{0.95x}{39.7} & 0.27 & \raptorspeed{0.98x}{44.1}
& 0.24 & \raptorspeed{0.94x}{25.3} & 0.24 & \raptorspeed{1.01x}{36.5}
& 0.33 & \raptorspeed{0.98x}{28.9} & 0.25 & \raptorspeed{0.95x}{31.9} \\
\specialrule{\lightrulewidth}{\aboverulesep}{0.2ex}
EAGLE
& 3.02 & 1.66x & 2.23 & 1.60x
& 2.47 & 1.57x & 2.01 & 1.40x
& 3.43 & 0.96x & 3.30 & 0.93x
& 2.34 & 1.54x & 2.69 & 1.38x \\
\specialrule{\lightrulewidth}{0.2ex}{\belowrulesep}
+SAGE
& 1.02 & \speeddown{0.64x}{61.4} & 1.06 & \speeddown{0.76x}{52.5}
& 1.15 & \speeddown{0.99x}{36.9} & 1.28 & \speeddown{0.89x}{36.4}
& 2.06 & \speedup{1.73x}{80.2} & 1.02 & \speeddown{0.70x}{24.7}
& 1.85 & \speeddown{1.11x}{27.9} & 1.35 & \speeddown{0.97x}{29.7} \\
+TALON
& 2.19 & \speedup{1.84x}{10.8} & 1.99 & \speedup{1.65x}{3.1}
& 1.93 & \speedup{1.83x}{16.6} & 2.30 & \speedup{1.60x}{14.3}
& 2.21 & \speedup{1.86x}{93.8} & 2.25 & \speedup{1.43x}{53.8}
& 2.20 & \speedup{1.76x}{14.3} & 2.15 & \speedup{1.71x}{23.9} \\
\rowcolor{raptorblue}
+Raptor
& 2.09 & \raptorspeed{2.01x}{21.1} & 1.98 & \raptorspeed{1.86x}{16.3}
& 2.03 & \raptorspeed{1.90x}{21.0} & 2.03 & \raptorspeed{1.88x}{34.3}
& 2.04 & \raptorspeed{2.02x}{110.4} & 2.07 & \raptorspeed{1.69x}{81.7}
& 2.12 & \raptorspeed{1.92x}{24.7} & 2.05 & \raptorspeed{1.90x}{37.7} \\
\specialrule{\lightrulewidth}{\aboverulesep}{0.2ex}
ViSpec
& 2.37 & 1.45x & 2.24 & 1.29x
& 2.36 & 1.38x & 2.27 & 1.32x
& 2.31 & 1.42x & 2.28 & 1.15x
& 2.41 & 1.39x & 2.32 & 1.34x \\
\specialrule{\lightrulewidth}{0.2ex}{\belowrulesep}
+SAGE
& 1.55 & \speedup{1.66x}{14.5} & 1.51 & \speedup{1.47x}{14.0}
& 1.54 & \speedup{1.55x}{12.3} & 1.58 & \speedup{1.63x}{23.5}
& 1.56 & \speedup{1.57x}{10.6} & 1.56 & \speedup{1.29x}{12.2}
& 1.56 & \speedup{1.61x}{15.8} & 1.55 & \speedup{1.54x}{14.9} \\
+TALON
& 2.19 & \speedup{1.84x}{26.9} & 2.08 & \speedup{1.55x}{20.2}
& 2.26 & \speedup{1.77x}{28.3} & 2.26 & \speedup{1.60x}{21.2}
& 2.30 & \speedup{1.93x}{35.9} & 2.20 & \speedup{1.43x}{24.3}
& 2.38 & \speedup{1.70x}{22.3} & 2.24 & \speedup{1.69x}{26.1} \\
\rowcolor{raptorblue}
+Raptor
& 2.26 & \raptorspeed{2.04x}{40.7} & 2.11 & \raptorspeed{1.85x}{43.4}
& 2.09 & \raptorspeed{1.87x}{35.5} & 2.11 & \raptorspeed{1.91x}{44.7}
& 2.17 & \raptorspeed{2.04x}{43.7} & 2.18 & \raptorspeed{1.68x}{46.1}
& 2.18 & \raptorspeed{1.89x}{36.0} & 2.16 & \raptorspeed{1.90x}{41.8} \\
\bottomrule
\end{tabular}%
}
\end{table*}

\section{Experiment}
\label{sec:exp}

\subsection{Experimental Setup}

\textbf{Hardware.} All experiments are conducted on two NVIDIA RTX 3090 GPUs with 24 GB memory each. Raptor operates exclusively during inference and introduces no additional training or parameter optimization. All methods are evaluated under the same hardware configuration.

\textbf{Models.} We evaluate Raptor on two representative open-source VLMs: LLaVA-1.6-Vicuna-7B and Qwen2.5-VL-7B-Instruct. Both models contain approximately 7 billion parameters and represent widely adopted architectures for multimodal understanding tasks. We integrate Raptor into different tree-structured speculative decoding frameworks, and Raptor neither modifies the underlying model architectures nor introduces additional trainable model parameters.

\textbf{Baseline.} We evaluate Raptor within three representative tree-structured speculative decoding frameworks: Medusa, EAGLE, and ViSpec. These frameworks employ different candidate-generation mechanisms, allowing us to examine whether rejection-aware computation allocation remains effective across distinct speculative decoding designs. For each framework, we compare its original configuration with versions equipped with Raptor or adapted tree-control policies from SAGE and TALON. The original configuration serves as the baseline for measuring the benefit of adaptation, while SAGE and TALON provide comparisons with existing adaptive strategies. Within each model–framework–benchmark combination, all strategies use the same target model, draft checkpoint, input examples, prompts, verification interface, and timing convention, while retaining their respective tree-construction policies. Raptor requires no additional model training or changes to the underlying model architectures. Implementation details of the adapted policies, node-budget conventions, and framework-specific execution are provided in Appendix B.

\textbf{Benchmarks.} We evaluate inference efficiency on fixed subsets of seven multimodal benchmarks: MM-Vet\cite{yu2024mm}, MME\cite{fu2026mme}, TextVQA\cite{singh2019towards}, ScienceQA (SQA)\cite{lu2022learn}, VizWiz\cite{gurari2018vizwiz}, COCO Captions\cite{chen2015microsoft}, and MMMU\cite{yue2024mmmu}. These benchmarks cover visual question answering, image captioning, and multimodal reasoning. Together, they span diverse image-conditioned generation settings, from recognizing scene content and reading text in images to answering knowledge-intensive questions. All strategies use identical selected examples and benchmark-specific prompt formats within each comparison, without task-specific fine-tuning. Dataset splits, subset sizes, and sampling procedures are detailed in Appendix B.2.

\textbf{Metrics.} All experiments use greedy decoding, with candidate tokens accepted according to the target model’s predictions. Raptor adapts candidate allocation and drafting effort while preserving the target-verification rule. We assess the resulting computational gains through accepted progress per verification round and measured generation throughput, using two efficiency metrics:
\begin{itemize}
\item \textbf{Average Acceptance Length} ($\tau$): The average maximum number of consecutive draft tokens accepted by the target model during each verification cycle, reflecting the quality of draft tree prediction.
\item \textbf{Speedup Ratio} (Spd.): The ratio of speculative-decoding throughput to autoregressive throughput on the same evaluation examples. Throughput is computed as the arithmetic mean of per-example generated tokens per second.
\end{itemize}
The measured generation interval includes target-model prefill, drafting, verification, cache updates, controller computation, and visual processing performed inside the generation function. Model loading and processing outside this interval are excluded, as detailed in Appendix B.3.

\begin{figure}[!t]
\centering
\includegraphics[width=0.6\columnwidth]{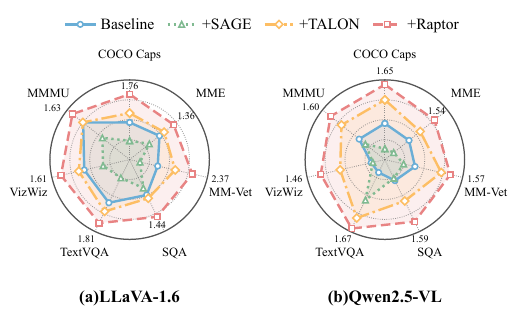}
\caption{Comparison between Raptor (ours) and existing efficient tree adaptation strategies on seven multimodal benchmarks. Speedups over autoregressive decoding are arithmetically averaged across Medusa, EAGLE, and ViSpec. Raptor achieves the highest average speedup on every benchmark for both models.}
\label{fig0}    
\end{figure}

\subsection{Acceleration Performance Comparison}
As shown in Table \ref{table1}, Raptor consistently improves generation throughput relative to the original speculative decoding configurations, achieving up to 2.39× average speedup and 3.21× peak speedup over autoregressive decoding. The magnitude of the improvement varies with the underlying drafter, reflecting differences in candidate quality, tree-construction cost, and the amount of serial expansion required by each framework. These gains extend across different candidate-generation mechanisms, supporting rejection-aware computation allocation as an effective adaptation strategy beyond a particular speculative decoding design.

Raptor also achieves the highest reported speedup among the evaluated adaptive strategies in all model–framework–benchmark configurations. As summarized in Figure~\ref{fig0}, its advantage persists when results are averaged across speculative decoding frameworks on each benchmark. Compared with confidence- or entropy-guided adaptation, Raptor combines target-side frontier coverage with accepted progress and redundant expansion to coordinate candidate allocation and drafting effort. The consistent throughput gains support the effectiveness of this joint feedback design across the evaluated multimodal generation settings.

The results further show that greater acceleration does not require longer accepted prefixes. For example, with EAGLE on Qwen2.5-VL, Raptor improves average speedup from 1.38× to 1.90× despite a decrease in acceptance length. This distinction highlights the limitation of treating acceptance length alone as a proxy for efficiency: additional accepted tokens must justify the computation spent constructing and verifying their candidates. Raptor addresses this trade-off by jointly regulating tree configuration and drafting effort, aligning speculative computation with useful decoding progress.

\definecolor{ablationgray}{RGB}{242,242,242}
\definecolor{ablationred}{RGB}{200,0,0}

\newcommand{\ablationdrop}[2]{%
  #1\hspace{0.06em}%
  \raisebox{-0.35ex}{%
    \scalebox{0.62}{%
      \normalfont%
      \textcolor{ablationred}{%
        \begingroup%
        \mathsurround=0pt%
        \ensuremath{\mathord{\downarrow}}%
        \kern-0.08em%
        #2\%%
        \endgroup%
      }%
    }%
  }%
}

\newsavebox{\ablationtreebox}
\newsavebox{\ablationvariantbox}
\newlength{\ablationcommonwidth}

\begin{table*}[t]
\centering
\renewcommand{\arraystretch}{1.08}
\footnotesize

\setlength{\tabcolsep}{3.2pt}
\sbox{\ablationtreebox}{%
\begin{tabular}{c|cccccc|cc}
\toprule
\multirow[c]{2}{*}[-\dimexpr
  \aboverulesep/2+\belowrulesep/2+\cmidrulewidth/2\relax
]{Configuration} &
\multicolumn{2}{c}{VizWiz} &
\multicolumn{2}{c}{SQA} &
\multicolumn{2}{c}{MMMU} &
\multicolumn{2}{|c}{Avg.} \\
\cmidrule(lr){2-3}
\cmidrule(lr){4-5}
\cmidrule(lr){6-7}
\cmidrule(lr){8-9}
& $\tau$ & Spd.
& $\tau$ & Spd.
& $\tau$ & Spd.
& $\tau$ & Spd. \\
\midrule
Raptor
& 2.18 & \textbf{1.68x}
& 2.11 & \textbf{1.91x}
& 2.18 & \textbf{1.89x}
& 2.16 & \textbf{1.83x} \\
\midrule
\rowcolor{ablationgray}
w/o Depth
& 2.23 & \ablationdrop{1.55x}{7.7}
& 2.19 & \ablationdrop{1.63x}{14.7}
& 2.20 & \ablationdrop{1.69x}{10.6}
& 2.21 & \ablationdrop{1.62x}{11.5} \\
w/o Width
& 2.09 & \ablationdrop{1.54x}{8.3}
& 2.04 & \ablationdrop{1.63x}{14.7}
& 2.05 & \ablationdrop{1.72x}{9.0}
& 2.06 & \ablationdrop{1.63x}{10.9} \\
\rowcolor{ablationgray}
w/o Budget
& 2.19 & \ablationdrop{1.64x}{2.4}
& 2.12 & \ablationdrop{1.83x}{4.2}
& 2.20 & \ablationdrop{1.82x}{3.7}
& 2.17 & \ablationdrop{1.76x}{3.8} \\
\bottomrule
\end{tabular}%
}

\setlength{\tabcolsep}{3.0pt}
\sbox{\ablationvariantbox}{%
\begin{tabular}{c|cccccc|cc}
\toprule
\multirow[c]{2}{*}[-\dimexpr
  \aboverulesep/2+\belowrulesep/2+\cmidrulewidth/2\relax
]{Variant} &
\multicolumn{2}{c}{VizWiz} &
\multicolumn{2}{c}{SQA} &
\multicolumn{2}{c}{MMMU} &
\multicolumn{2}{|c}{Avg.} \\
\cmidrule(lr){2-3}
\cmidrule(lr){4-5}
\cmidrule(lr){6-7}
\cmidrule(lr){8-9}
& $\tau$ & Spd.
& $\tau$ & Spd.
& $\tau$ & Spd.
& $\tau$ & Spd. \\
\midrule
Raptor
& 2.18 & \textbf{1.68x}
& 2.11 & \textbf{1.91x}
& 2.18 & \textbf{1.89x}
& 2.16 & \textbf{1.83x} \\
\midrule
\rowcolor{ablationgray}
w/o EMA
& 2.29 & \ablationdrop{1.55x}{7.7}
& 2.31 & \ablationdrop{1.62x}{15.2}
& 2.11 & \ablationdrop{1.71x}{9.5}
& 2.24 & \ablationdrop{1.63x}{10.9} \\
w/o Momentum
& 2.26 & \ablationdrop{1.52x}{9.5}
& 2.47 & \ablationdrop{1.64x}{14.1}
& 2.24 & \ablationdrop{1.77x}{6.3}
& 2.32 & \ablationdrop{1.64x}{10.4} \\
\rowcolor{ablationgray}
w/o Both
& 2.16 & \ablationdrop{1.39x}{17.3}
& 2.11 & \ablationdrop{1.58x}{17.3}
& 2.13 & \ablationdrop{1.64x}{13.2}
& 2.13 & \ablationdrop{1.54x}{15.8} \\
\bottomrule
\end{tabular}%
}

\setlength{\ablationcommonwidth}{\wd\ablationtreebox}
\ifdim\wd\ablationvariantbox>\ablationcommonwidth
  \setlength{\ablationcommonwidth}{\wd\ablationvariantbox}
\fi

\begin{minipage}[t]{0.49\textwidth}
\vspace{0pt}
\centering
\caption{Ablation study of the components of adaptive budget regulation.}
\label{table2}
\resizebox{\linewidth}{!}{%
  \makebox[\ablationcommonwidth][c]{%
    \usebox{\ablationtreebox}%
  }%
}
\end{minipage}%
\hfill
\begin{minipage}[t]{0.49\textwidth}
\vspace{0pt}
\centering
\caption{Ablation study on EMA smoothing and momentum update.}
\label{table3}
\resizebox{\linewidth}{!}{%
  \makebox[\ablationcommonwidth][c]{%
    \usebox{\ablationvariantbox}%
  }%
}
\end{minipage}
\end{table*}

\begin{figure*}[!t]
\centering
\includegraphics[width=1.0\textwidth]{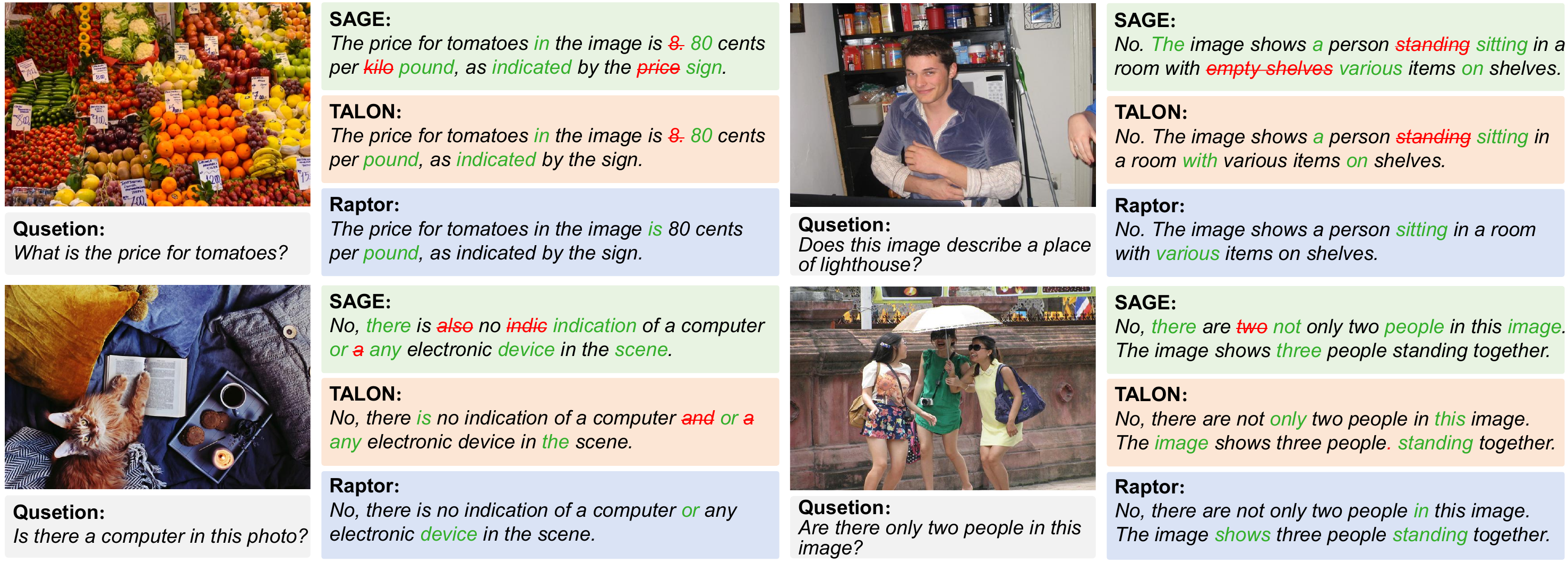}
\caption{\textbf{Comparison of generation traces for SAGE, TALON, and Raptor.} Each example presents the input image, question, and corresponding decoding traces. Black text denotes accepted draft tokens, green text denotes target-generated tokens, and red strikethrough text denotes rejected draft tokens. In these examples, Raptor adapts candidate exploration using rejection feedback, reducing rejections and corrections during generation.}
\label{fig6}
\end{figure*}

\subsection{Ablation Study}

To examine the contribution of individual components, we conduct ablation studies on joint configuration adaptation and temporal smoothing. Table \ref{table2} evaluates depth, width, and node-budget adaptation by fixing one variable at a time while retaining adaptation of the others. Table \ref{table3} examines EMA smoothing and momentum updates by removing each component separately and both together.

\textbf{Joint Configuration Adaptation.} We ablate joint configuration adaptation by fixing depth, width, or node budget while keeping the other variables adaptive. As shown in Table 2, each ablation reduces speedup across the evaluated benchmarks, with larger average losses from fixing depth or width. Depth and width balance path extension against candidate exploration, while the node budget controls total speculative computation. Because these variables constrain one another through the feasible tree capacity, changing one dimension also changes the computational value of the others. These results support jointly adapting tree depth, branching width, and node budget rather than optimizing each dimension independently.

The acceptance results further illustrate why these decisions should be coordinated. On SQA, fixing depth increases acceptance length from 2.11 to 2.19 but reduces speedup from 1.91× to 1.63×. Fixing the node budget likewise maintains or slightly increases acceptance length across all evaluated benchmarks while lowering throughput. These findings indicate that additional accepted tokens do not necessarily compensate for the computation required to obtain them. Joint configuration adaptation achieves a more favorable balance between accepted progress and execution cost in the evaluated setting.

\textbf{EMA Smoothing and Momentum Updates.} We further evaluate the impact of EMA smoothing and momentum updates on feedback regulation. EMA stabilizes the verification statistics used to compute control pressure, whereas momentum preserves the recent update direction across decoding rounds. This separation enables Raptor to respond to persistent changes in speculative behavior without overreacting to individual verification outcomes. As shown in Table \ref{table3}, removing either component reduces speedup across the evaluated benchmarks, while removing both yields the lowest average speedup, confirming their complementary roles in efficient adaptation.

The acceptance results show that longer accepted prefixes do not necessarily yield higher throughput, highlighting the importance of how verification feedback is translated into configuration changes. The complete Raptor controller combines smoothed observations with momentum-guided updates to moderate its response to round-to-round variation. Its consistent speedup advantage over the ablated variants supports retaining both components for effective feedback regulation and efficient speculative computation.

\begin{figure*}[t]
\centering
\includegraphics[width=1.0\textwidth]{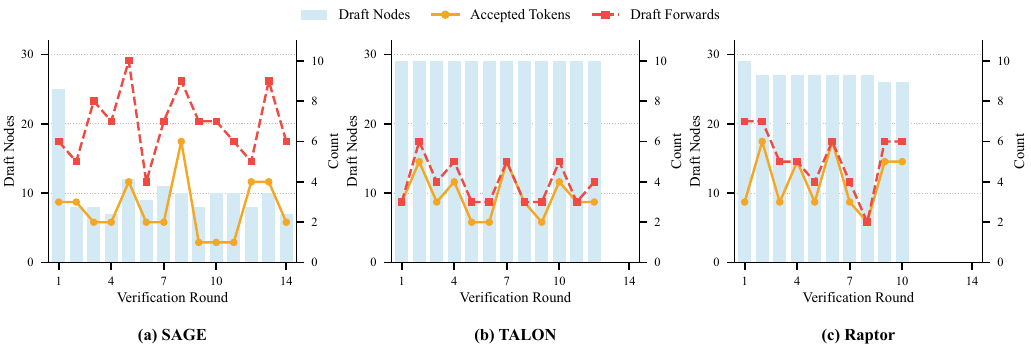}
\caption{\textbf{Per-round decoding behavior under different tree adaptation strategies.} 
Light-blue bars show the number of non-root draft nodes submitted for verification (left axis). The orange solid and red dashed lines indicate accepted draft tokens and actual draft forward calls, respectively (right axis). Raptor achieves longer accepted prefixes on average and requires fewer verification rounds.}
\label{fig7}
\end{figure*}

\subsection{Qualitative Analysis}
The generation traces in Figure~\ref{fig6} illustrate how speculative computation translates into accepted progress. In the displayed examples, Raptor produces fewer rejected draft continuations, reducing computation on candidates excluded from the committed output. This reflects its joint consideration of frontier coverage and redundant expansion: exploration follows verification outcomes, while the same number of submitted nodes can incur different drafting costs depending on how the tree is expanded. This distinction is important for VLM inference, as Raptor directs speculation toward accepted progress and improves the efficiency of the draft-and-verify process.

The per-round behavior further highlights the distinction between tree size and execution cost. This distinction is important for VLM inference because the same number of submitted nodes can incur different drafting costs depending on how the tree is expanded. As shown in Figure~\ref{fig7}, SAGE submits fewer verification nodes in many rounds but incurs substantial drafting effort, whereas Raptor maintains productive acceptance with fewer drafting forwards in much of the trace. Compared with TALON, Raptor completes the generation in fewer verification rounds without increasing the number of submitted nodes. Together, these observations illustrate the benefit of coordinating candidate allocation and drafting effort: reducing tree size alone is insufficient, and speculative computation must translate into useful progress to reduce repeated draft-and-verify execution.

\section{Conclusion}
\label{sec:conclusion}

We presented Raptor, a training-free framework for rejection-aware computation allocation in VLM speculative decoding. By combining accepted progress, frontier coverage, and redundant expansion, Raptor jointly regulates tree depth, branching width, and node budget through bounded, momentum-smoothed updates. Rejection-aware execution control further regulates drafting effort by distinguishing shallow rejection from path exhaustion. Experiments demonstrate consistent throughput gains over original SD configurations and the evaluated adaptive tree strategies, while ablations support the complementary roles of joint configuration adaptation and temporal smoothing. These findings highlight the value of coordinating candidate coverage and speculative effort through verification feedback, demonstrating Raptor’s effectiveness in improving VLM inference efficiency.

\bibliographystyle{unsrt}  
\bibliography{references}  






\clearpage
\setcounter{page}{1}

\appendix
\renewcommand{\thesection}{\Alph{section}}

\section{Algorithmic and Implementation Details}
\label{sec:algorithm}
\subsection{Decoding Procedure}

Algorithm 1 summarizes the closed-loop decoding procedure of Raptor. At the beginning of round \(t\), the controller maintains a tree configuration ${{\bf{B}}_t} = {\left[ {{D_t}{\text{ }}{W_t}{\text{ }}{N_t}} \right]^{\rm{T}}}$, which specifies the drafting depth, branching width, and node budget, respectively. The drafter constructs a candidate tree under the current configuration, and the target model verifies its candidate paths in parallel. Raptor extracts acceptance length, frontier gap, and deep waste from the verification results and produces the next configuration through the smoothing, pressure mapping, and momentum updates defined in the main text. The tree configuration determines the candidate exploration range, while the execution rules in Section A.3 further constrain the actual expansion process.

We use \(N_t\) to denote the total node budget, including the verification root. The implementation uses the corresponding non-root draft budget $n_t=N_t-1$. Each question starts from \((D_1,W_1,N_1)=(6,5,30)\). The continuous configuration is initialized to the corresponding discrete values, the update velocity and consecutive shallow-rejection counter are initialized to zero, and the additional expansion limit is initially unset. Each EMA is initialized with the first observation of its signal. These states are reset independently for every question.

The acceptance length \(a_t\) counts accepted non-root draft tokens. Under the verification interface, the root and the accepted draft prefix form the sequence committed in the current round, yielding an advancement of \(a_t+1\) tokens before termination handling. After verification, the caches associated with the accepted path are updated so that subsequent generation continues from the committed prefix. 
The current implementation checks EOS and the generation-length condition before committing the accepted sequence. Consequently, the terminal round may retain tokens committed after the first EOS within that round. Experimental token counts follow the implementation output; the timing and metric conventions are specified in Appendix B.

\subsection{Control Matrix Settings}

Raptor uses a fixed matrix to transform rejection pressures into configuration increments. Let \(\mathbf P_t=[P_t^a \text{ } P_t^g \text{ } P_t^w]^\top\) denote the acceptance, coverage, and waste pressures. For the internal configuration \(\mathbf B_t=[D_t \text{ } W_t \text{ } N_t]^\top\), the transformation is

\[
\Delta\mathbf B_t=\mathbf W \cdot \mathbf P_t,\qquad
\mathbf W=
\begin{bmatrix}
0.55&0&-0.75\\
0.20&0.90&-0.15\\
0.70&0.35&-0.65
\end{bmatrix}.
\]

\begin{algorithm}[tb]
\caption{Rejection-aware Adaptive Tree Decoding}
\label{alg:algorithm}
\textbf{Input}: Initial tree configuration $\textbf{B}_0$,\\
\phantom{\textbf{Input:}} Maximum generated length L.\\
\textbf{Parameter}: Momentum coefficient $\mu$,\\
\phantom{\textbf{Parameter:}} Update step size $\eta$,\\  
\phantom{\textbf{Parameter:}} Pressure transformation matrix \textbf{W}.\\
\textbf{Output}: Generated token sequence \textbf{Y}. 

\begin{algorithmic}[1] 
\STATE $\textbf{B}_1, \textbf{ B}_1^{\left( f \right)} \gets {\textbf{B}_0}$, 
        $\mathbf{ v}_1 \gets 0, \text{ } t \gets 1, \mathbf{ Y} \gets \varnothing$.
\WHILE{$(|\mathbf{Y}| < L \land
        \mathrm{EOS} \notin \mathbf{Y})$}
    \STATE $\mathcal{T}_t \gets
            \operatorname{Draft}(\mathbf{B}_t)$.
    \STATE $(\textbf{C}_t,a_t,g_t,w_t) \gets
           \operatorname{Verify}(\mathcal{T}_t)$.
    \STATE $\mathbf{C}_t \gets
           \operatorname{Take}_{L-|\mathbf{Y}|}
           (\operatorname{PrefixThroughEOS}(\mathbf{C}_t))$.
    \STATE $\mathbf{Y} \gets \mathbf{Y} \Vert \mathbf{C}_t$.
    \STATE Convert rejection statistics $(a_t,g_t,w_t)$ into bounded pressures $\textbf{P}_t$.
    \STATE $\Delta {\textbf{B}_t} = \textbf{W} \cdot {\textbf{P}_t}$.
    \STATE ${\textbf{v}_t} \gets \mu {\textbf{v}_{t - 1}} + \left( {1 - \mu } \right)\Delta {\textbf{B}_t}$.
    \STATE $\textbf{B}_{t + 1}^{\left( f \right)} \gets \textbf{B}_t^{\left( f \right)} + \eta {\textbf{v}_t}$.
    \STATE $\textbf{B}_{t + 1} \gets \text{clip}\left( {\text{round}\left( {\textbf{B}_{t+1}^{\left( f \right)}} \right)} \right)$.
    \STATE $t \gets t+1$
\ENDWHILE
\STATE \textbf{return} generated sequence \textbf{Y}
\end{algorithmic}
\end{algorithm}

The rows correspond to depth, width, and the non-root node budget, while the columns correspond to the three pressure channels. The matrix remains fixed during decoding, whereas the tree configuration changes with verification feedback.

The coefficients were empirically calibrated using separate examples from multiple benchmarks. The calibration examples do not overlap with those used in the final efficiency experiments, separating coefficient selection from final result reporting at the example level. Examples from multiple benchmarks provided observations of controller responses under different task contents and generation behaviors, from which the shared matrix above was established. This procedure sets fixed controller hyperparameters and does not train the target or draft model weights.

The coefficients should be interpreted together with the pressure definitions. When the smoothed acceptance length falls below its reference operating point, the acceptance pressure is positive and contributes positive adjustments to all three configuration components. The coverage pressure primarily increases width and provides additional budget for candidate exploration; its direct coefficient for depth is zero. The waste pressure contributes negative adjustments to all three components, with larger coefficient magnitudes for depth and budget. Thus, expansion driven by insufficient advancement is jointly constrained by evidence of redundant computation, and the final direction of each configuration update depends on all three pressures.

These coefficients specify continuous update directions whose effects also depend on momentum, step size, and configuration bounds. The implementation applies twice the base step size to the budget channel. Continuous configurations are constrained by their ranges and capacity before rounding, followed by an integer-capacity check. We denote the transformation matrix by \(\mathbf W\) to distinguish it from the branching-width variable. The separation between calibration and evaluation examples avoids selecting coefficients directly on the final evaluation instances; it does not by itself establish performance on untested tasks or models.

\subsection{Expansion Control and Recovery}

Raptor maintains an execution state alongside the \(D/W/N\) controller to reduce serial expansion with limited benefit. Let \(c_t\) denote the requested additional expansion limit in round \(t\), where \(c_t=\varnothing\) indicates that no additional limit is imposed. This state is updated from actual verification feedback and is constrained by the current tree configuration and drafting architecture.

Raptor distinguishes genuine shallow rejection from path exhaustion. A round is classified as a genuine shallow rejection, denoted by \(s_t=1\), if the selected path still contains unaccepted candidates, the last accepted node lies at no more than half of the maximum tree depth, and at least two levels separate these depths. Path exhaustion, denoted by \(\ell_t=1\), occurs when all valid candidates on the selected path have been accepted. Since acceptance beyond that path remains unobserved, this event is treated separately from genuine shallow rejection. Exhaustion refers to the selected path and does not require its leaf to reach the maximum depth of the entire tree.

The execution layer also retains a one-level reduction rule for consecutive shallow rejections. If at least two preceding rounds exhibit genuine shallow rejection and the candidate capacity after a one-level reduction can still accommodate the current draft budget, the originally planned expansion is reduced by one level before tree construction. EAGLE and ViSpec check capacity using the current width and remaining expansion levels, while Medusa performs the corresponding check using its available prediction heads. The reduction is applied when structurally feasible and affects only the current tree construction. The stored discrete depth is restored afterward, while its continuous state remains governed by the core controller.

Before tree construction, Raptor clips the requested limit to the current depth and the drafter’s executable range, yielding the effective limit \(c_t^{\mathrm{eff}}\). No additional restriction is imposed when the requested limit is unset. For EAGLE and ViSpec, reaching the effective limit stops expansion only if the candidate pool contains enough non-root candidates to satisfy the current budget. Otherwise, expansion continues until the budget is satisfied or the executable ceiling is reached. The effective limit therefore specifies a preferred stopping point, with deeper expansion permitted when candidates are insufficient, subject to the current configuration and architectural constraints.

After verification, the core controller first produces \(D_{t+1}\). Defining \(\bar D_t=\min(D_t,D_{t+1})\), the requested limit for the next round is

\[
c_{t+1}=
\begin{cases}
\min(2,\bar D_t), & s_t=1,\\
\min(c_t^{\mathrm{eff}}+1,\bar D_t),
& \ell_t=1,\ c_t^{\mathrm{eff}}\ne\varnothing,\\
\varnothing, & \text{otherwise}.
\end{cases}
\]

A single genuine shallow rejection is sufficient to request a two-expansion limit for the next round. When the selected path is exhausted and an effective limit exists, the limit increases by one, allowing progressive recovery through \(2\rightarrow3\rightarrow4\rightarrow\cdots\). Recovery is bounded jointly by the current and next-round depths.

Recovery is computed from the effective limit used in the current round. Even if budget fallback causes more expansions to be executed, the next limit still increments from \(c_t^{\mathrm{eff}}\). The implementation applies this update whenever path exhaustion occurs under an effective limit, regardless of whether that limit actually reduced expansion in the current round. Rejection feedback therefore regulates both the next tree configuration and its execution range, allowing deeper exploration to recover progressively as generation behavior changes.

\subsection{Drafter Integration}

Raptor integrates with Medusa, EAGLE, and ViSpec through a shared tree representation and verification-feedback interface. Candidate tokens, retrieval-path indices, an ancestor attention mask, and node positions jointly specify the verification tree. After verification, the same controller receives rejection signals and updates the next \(D/W/N\) configuration and execution state.

EAGLE expands candidate paths through serial draft-model forwards, so the additional limit applies directly to its tree-construction loop. ViSpec applies the same execution constraint to its vision-aware drafter while retaining its visual adaptation and input processing. Both methods continue expansion when the candidate pool is insufficient and retain their original candidate-ranking and tree-assembly procedures. Draft-prefix computation and subsequent expansion are counted separately; two serial expansions typically require one prefix forward and two expansion forwards.

Medusa uses different prediction heads to provide candidates for different future positions. The first head generates the initial candidates, and subsequent heads support further expansion; two expansions therefore typically correspond to three prediction heads. When an effective limit exists, Raptor determines the number of heads to compute from both that limit and the number required to fill the candidate budget. Additional heads may be evaluated if the limited set provides insufficient candidates, subject to the current configuration and the model’s finite head count. Without an additional limit, all heads available in the current round are used.

A single Medusa draft-module call can compute multiple prediction heads. We therefore report its head computations separately from the serial forward counts of EAGLE and ViSpec. Final efficiency across architectures is measured by token throughput under the same timing convention. Model weights, cache implementations, datasets, and evaluation settings are specified in Appendix B.

\section{Experimental Details}
\label{sec: Experimatal Details}

\subsection{Models and Configuration}

We evaluate Raptor on LLaVA-1.6-Vicuna-7B and Qwen2.5-VL-7B-Instruct, integrating it with Medusa, EAGLE, and ViSpec. Each method is evaluated on seven multimodal benchmarks, yielding \(2\times3\times7=42\) combinations for each adaptive strategy. Experiments use two NVIDIA RTX 3090 GPUs with 24 GB of memory each. Combinations are executed sequentially by the launch scripts, and models are loaded with automatic device mapping.

All strategies retain the target model, draft checkpoint, task arguments, and prompt settings of the corresponding experimental profile. Models are loaded with automatic precision selection, and the decoding temperature is zero. The initialization seed is 42. Evaluators retain their existing warmup and per-answer seed settings, and warmup time is excluded from the measured generation time of evaluation examples.

Raptor starts each question with \(D=6\), \(W=5\), and a total node budget of \(N=30\), corresponding to 29 non-root draft nodes internally. TALON uses a total budget of 30, five root candidates, and a relative confidence threshold of 0.03. SAGE also uses a maximum of 30 nodes while retaining its fixed initial tree and subsequent adaptive rules. Node budgets include the verification root, and each method determines the actual tree structure according to its own policy.

Weight-loading information is saved with the runtime configuration. The EAGLE and ViSpec entry points terminate when required draft parameters are missing. Medusa retains the historical loading convention, allowing parameters absent from the checkpoint to keep their initialized values while recording missing keys and the initialization seed. Weight-loading conditions are held consistent across strategies using the same drafter.

\subsection{Data and Sampling}

We evaluate on MM-Vet, MME, TextVQA, ScienceQA, VizWiz, COCO Caption, and MMMU. All strategies reuse the project’s existing data loaders and prompt-construction procedures. Before inference, selected question IDs are checked against historical sample records to ensure that each experimental combination uses the same inputs.

MM-Vet uses the test split of the local dataset, with 218 question IDs in the historical complete run. For MME, the loaded records are shuffled with seed 42, and the first 100 records are selected. These records contain a repeated ID, resulting in 99 unique records after the original evaluator organizes outputs by question ID. The sampled input order is preserved, and input-record counts and final unique IDs are checked separately.

TextVQA uses the TextVQA 0.5.1 test data, and VizWiz uses the local test data. Both datasets are shuffled with seed 42 before selecting the first 100 records. Images are loaded after metadata selection, preserving the original sampling order and image-serialization procedure while reducing host-memory usage from unselected images.

ScienceQA first retains questions with images and then samples up to 100 questions from the test split using a fixed seed. It uses the zero-shot QCM-ALE prompt format. COCO Caption constructs image records from the MS COCO 2017 validation split, and we select 100 images for evaluation after shuffling with a fixed seed. MMMU reuses the existing loader to collect all subjects from the test split, shuffles the records with seed 42, and selects 100 examples. Each benchmark retains its evaluator-specific prompt format, with identical input and prompting settings across strategies within that benchmark.

\subsection{Timing and Metrics}

All strategies use the same full-generation timing interface. Timing starts immediately before generation and ends after the function returns, with all visible CUDA devices synchronized at both boundaries. The interval includes target-model prefill, drafting, verification, cache updates, controller computation, and visual processing performed inside the generation function. Model loading, prompt construction, external warmup, answer decoding, and result writing are excluded.

Let \(G_q\) and \(T_q\) denote the generated-token count and generation time of evaluation record \(q\). Speculative-decoding throughput is the arithmetic mean of per-example throughputs, and speedup is measured relative to the corresponding AR throughput.

\subsection{Baseline Methods}

\textbf{TALON.} We adapt TALON’s confidence-aware tree policy to the three existing drafters. The first level selects up to five candidates from the root distribution. At subsequent levels, parent-path log probabilities are added to conditional draft log probabilities, and a relative threshold is applied jointly across all parents and the full vocabulary. Candidates with scores at least equal to the level maximum plus \(\log(0.03)\) are retained. If they exceed the remaining budget, the highest-scoring candidates are selected. No additional per-parent top-k prefilter is applied at subsequent levels, and expansion stops when the total node budget is reached.

EAGLE and ViSpec compute subsequent conditional distributions with their respective draft modules. Medusa provides candidates through its available prediction heads and is bounded by their finite number. Heads are evaluated as needed and are not recycled to fabricate additional depth. Constructed trees are converted to the shared candidate, retrieval-path, attention-mask, and position representation for verification.

\textbf{SAGE.} We retain SAGE’s top-k normalized-entropy confidence, confidence-based depth and width adaptation, hierarchical width decay, cumulative-path-probability threshold, and acceptance-history updates. We use \(k=10\), depth bounds of 3–8, width bounds of 2–10, a history window of 10, and lower and upper acceptance thresholds of 2 and 3. The maximum node count is set to 30 to match the budget constraint of our experiments. The first round uses the fixed tree template corresponding to the paper’s appendix, followed by dynamic construction.

Confidence is retained during drafting and committed after verification together with the acceptance-history update, informing the next tree. Acceptance history changes the effective maximum depth by one step within its configured bounds. Construction preserves probability-ranked depth-first node admission. The runtime batches sibling computations while sharing ancestor prefixes and isolating attention between siblings. Computation for batched siblings that are not ultimately admitted to the tree is included in measured runtime.

Fixed reproduction conventions resolve discretization details not uniquely specified by the paper. These include rounding depth and width interpolations half up, flooring local widths, and interpreting parent probability as cumulative path probability. Medusa depth is bounded by its actual prediction heads, with path scores formed from independent-head probabilities. Each drafter retains its original visual processing, and no additional SpecVLM visual-token pruning from SAGE’s original experiments is introduced.

\subsection{Fairness and Reproducibility}

Comparisons within each model–drafter–benchmark combination use the same target model, draft checkpoint, example IDs, prompts, verification interface, and timing convention. Speedups are evaluated against the same AR baseline. Each method retains its own initialization and tree policy, avoiding changes to its design solely to impose a common tree shape. Raptor regulates \(D/W/N\) and execution range through rejection feedback, TALON selects candidates through relative confidence, and SAGE adapts trees using entropy and acceptance history.

Experiments save runtime arguments, weight-loading information, and output records. Inputs are checked against historical records before inference, and SD and AR question IDs are checked again during evaluation to ensure that speedups correspond to the same examples. Resumption validates the complete sample set before skipping completed records. Completed examples retain their recorded times, while remaining examples are measured in the resumed process. Inconsistent configurations or mismatched sample sets terminate execution.

The full-suite scripts execute experimental combinations in a fixed order. We report the throughput and speedup observed under this procedure and do not characterize these runs as randomized-order or repeated measurements.

\begin{figure*}[!t]
\centering
\includegraphics[width=1.0\textwidth]{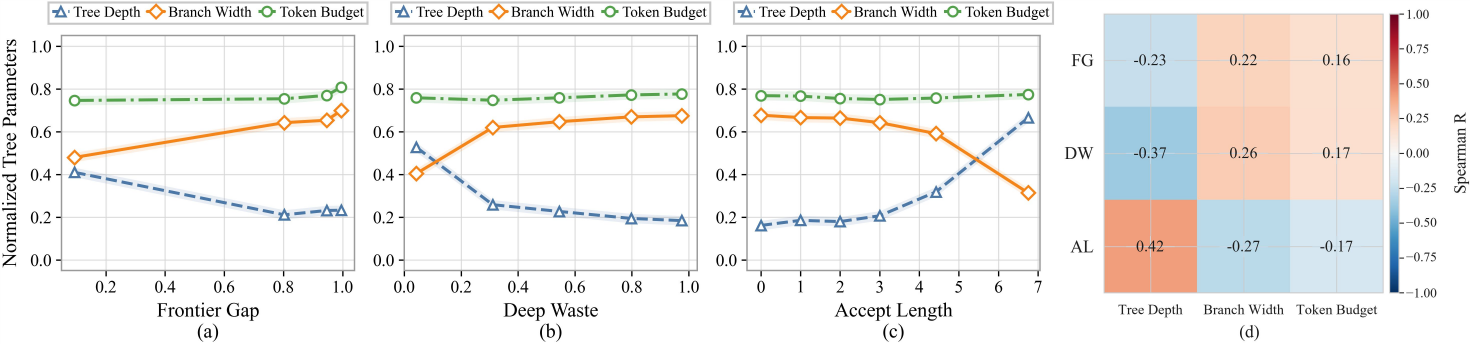}
\caption{\textbf{Relationship between rejection-aware signals and subsequent proxy reference configurations.} (a)–(c) show trends in normalized reference depth, branching width, and node budget conditioned on each signal. (d) reports correlations between the signals and reference configuration dimensions. FG, DW, and AL denote frontier gap, deep waste, and acceptance length, respectively.}
\label{fig2}    
\end{figure*}

\section{Motivation}
\label{sec:moti}

In VLM speculative decoding, increasing candidate coverage also introduces additional drafting and verification work over multimodal contexts. We therefore examine whether verification feedback can help distinguish where speculative computation is useful. Using fixed-tree decoding traces, we relate the current round’s acceptance length, frontier gap, and deep waste to proxy reference configurations for the subsequent round. These references are constructed from predefined acceptance and coverage requirements and are not measured runtime optima. Acceptance length describes achieved progress, frontier gap measures target probability mass missing from the local candidate frontier, and deep waste summarizes expansion beyond the accepted prefix. This analysis illustrates allocation trade-offs rather than establishing an optimal tree-selection rule.

\subsection{Observation 1: Rejection signals reflect distinct allocation trends}

We first examine how rejection-aware signals observed in the current decoding round relate to reference tree configurations in the next round. As shown in Figure~\ref{fig2}(a)–(c), larger frontier gaps are associated with wider and shallower reference trees, whereas longer accepted prefixes are associated with deeper and narrower configurations. Greater deep waste is also associated with reduced reference depth, highlighting the relevance of expansion beyond the accepted prefix to subsequent allocation. These trends suggest that acceptance length alone is insufficient to guide subsequent computation allocation, motivating a feedback mechanism that combines complementary rejection signals.

\subsection{Observation 2: Rejection signals reveal distinct optimization dimensions}

Beyond individual trends, we further examine the correlations between rejection signals and reference configuration dimensions. As shown in Figure~\ref{fig2}(d), acceptance length correlates positively with depth and negatively with width, while frontier gap and deep waste exhibit the opposite depth–width pattern. Their correlations with node budget are comparatively modest. In particular, the positive marginal association between deep waste and reference budget should not be interpreted as a rule to increase computation when waste rises. The next allocation depends on the joint verification state and the construction of the reference configuration.

\begin{figure}[h]
\centering
\includegraphics[width=0.5\textwidth]{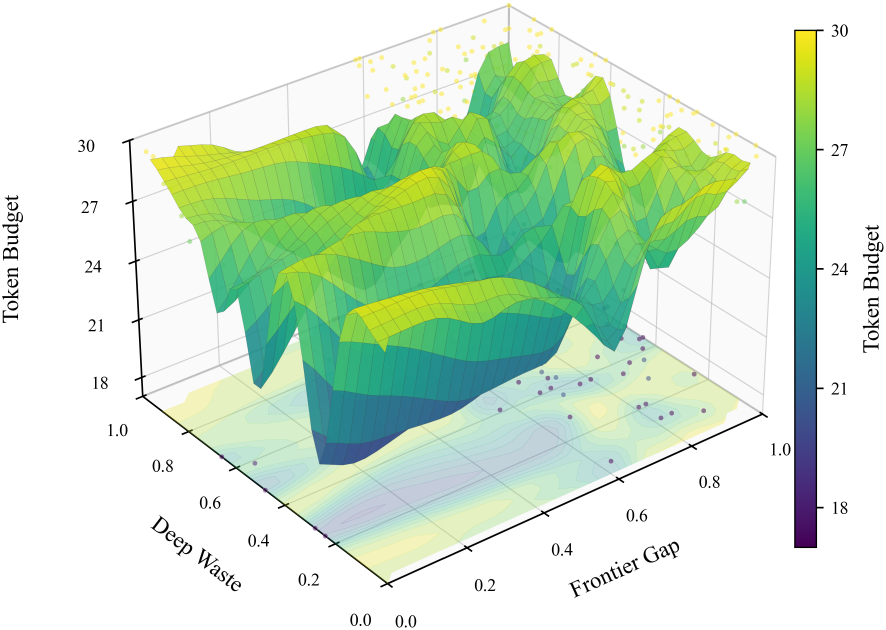}
\caption{Variation in proxy reference node budgets across combinations of frontier gap and deep waste.}
\label{fig3}
\end{figure}

\subsection{Observation 3: Drafting token budget varies under diverse rejection states}

While tree depth and width determine speculative structure, the drafting budget controls the overall allocation of speculative computation. To examine how budget requirements vary with rejection feedback, we analyze drafting budgets under different rejection states characterized by frontier gap and deep waste. As shown in Figure~\ref{fig3}, reference budgets vary across combinations of the two signals, highlighting the need to consider candidate coverage and redundant expansion jointly. Insufficient coverage motivates additional candidate exploration, whereas excessive deep expansion calls for restraining unproductive computation. These competing demands motivate joint feedback regulation of the drafting budget.

\end{document}